\documentclass{article}
\usepackage[nonatbib, preprint]{neurips_2023}
\usepackage[utf8]{inputenc} 
\usepackage[T1]{fontenc}    
\usepackage{url}            
\usepackage{booktabs}       
\usepackage{amsfonts}       
\usepackage{nicefrac}       
\usepackage{microtype}      
\usepackage{xcolor}         
\usepackage{graphicx}
\usepackage{comment}
\usepackage{multicol}
\usepackage{color}
\usepackage{tabularray}
\usepackage{rotating}
\usepackage{adjustbox}
\usepackage{multirow}
\usepackage{makecell}
\usepackage{caption}
\usepackage{changepage}
\usepackage[english]{babel}
\usepackage{caption}
\usepackage{hanging}

\title{\fontsize{16pt}{16pt}\selectfont KoMultiText: Large-Scale Korean Text Dataset for Classifying Biased Speech in Real-World Online Services}

\author{Dasol Choi$^{1}$ \; Jooyoung Song$^{2}$ \; Eunsun Lee$^{1}$ \; Jinwoo Seo$^3$ \; Heejune Park$^4$ \; Dongbin Na$^{5}$\thanks{Correspondence to dongbinna@postech.ac.kr} \\ $^1$Kyunghee University \; $^2$Hongik University \; $^3$Catholic University \; $^4$Dankook University \; $^5$POSTECH}

\begin{document}

\maketitle

\begin{abstract}

With the growth of online services, the need for advanced text classification algorithms, such as sentiment analysis and biased text detection, has become increasingly evident.
The anonymous nature of online services often leads to the presence of biased and harmful language, posing challenges to maintaining the health of online communities.
This phenomenon is especially relevant in South Korea, where large-scale hate speech detection algorithms have not yet been broadly explored.
In this paper, we introduce  "KoMultiText", a new comprehensive, large-scale dataset collected from a well-known South Korean SNS platform.
Our proposed dataset provides annotations including (1) \textbf{Preferences}, (2) \textbf{Profanities}, and (3) \textbf{Nine types of Bias} for the text samples, enabling multi-task learning for simultaneous classification of user-generated texts. 
Leveraging state-of-the-art BERT-based language models, our approach surpasses human-level accuracy across diverse classification tasks, as measured by various metrics.
Beyond academic contributions, our work can provide practical solutions for real-world hate speech and bias mitigation, contributing directly to the improvement of online community health. 
Our work provides a robust foundation for future research aiming to improve the quality of online discourse and foster societal well-being.
All source codes and datasets are publicly accessible at \textcolor{blue}{\url{https://github.com/Dasol-Choi/KoMultiText}}.

\end{abstract}

\section{Introduction}

Various online platforms including social network services (SNS) have become the main communication channels in modern societies.
These platforms allow users to freely express opinions and interact globally. However, this freedom can also lead to the spread of biased or hateful speech~\cite{hs_cyber, hs_socialmedia, Spread_hs, beep}. Anonymity features in these services sometimes result in undesirable consequences, such as adverse effects on celebrities and individuals leading to severe mental trauma~\cite{beep}.

To address this issue, several online platforms in South Korea have implemented guidelines and automated detection methods for harmful speech~\cite{autodetection_sn, autodetection_hs, autodetection_fb, autodetection_cb}
However, rule-based detection algorithms might result in both a lot of false positives and false negatives due to their inability to capture the nuances of human language.
To tackle this problem, we provide a comprehensive, large-scale dataset collected from a well-known South Korean SNS platform.
Our dataset is designed to advance the field of text classification. Moreover, our dataset can facilitate the automated identification and mitigation of hate speech and various biases.
Compared to previous studies that only address multi-class classification~\cite{beep, mulclass_1, kohs_dataset}, our dataset offers a more nuanced multi-label classification scheme.
Our dataset includes extensive annotations for \textit{User Preferences}, \textit{Profanities}, and \textit{Nine distinct types of Biases per text, providing a useful granularity for more accurate comment analysis.}

\begin{figure*}[htp]
    \centering    \includegraphics[width=0.75\textwidth]{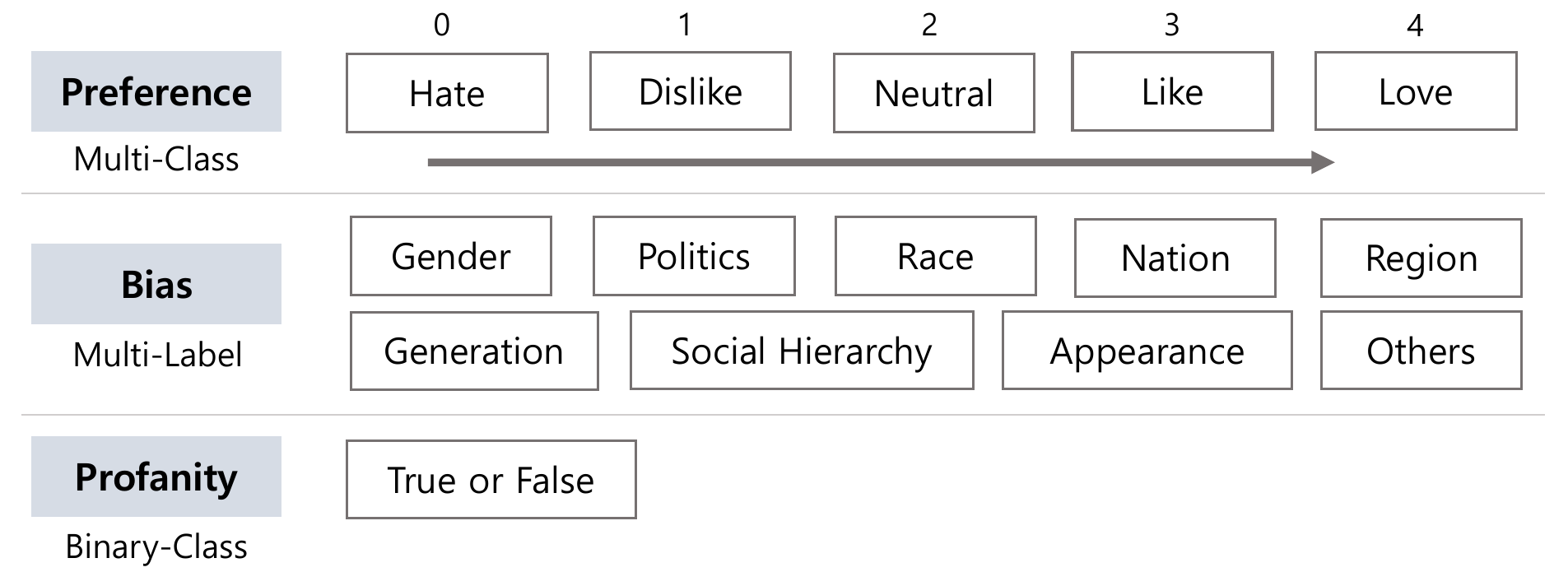}
    \caption{An illustration of our large-scale dataset. Each text in this dataset contains three kinds of annotations: (1) Preference label denoting the magnitude of the writer's preference (multi-class classification). (2) Bias labels (multi-label classification). (3) Profanities (binary classification).}
    \label{fig:dataset_configuration}
\end{figure*}

To validate the usefulness of our proposed dataset, we leverage state-of-the-art transformer architectures, specifically KR-BERT, KoBigBird, RoBERTa, and KoELECTRA~\cite{krbert, bigbird, roberta, electra}. 
We fine-tune the BERT-like pre-trained models on our proposed dataset to solve the multi-task problems simultaneously (1) Detecting one of five multi-class preferences, (2) Identifying the presence or absence of profanity through binary classification, and (3) Classifying multiple types of biases using multi-label annotations.
This multi-task approach can be a comprehensive solution for the moderation of text contents with various nuances, surpassing traditional methods in detection performance~\cite{multitask_recurrent, overview_mtl, KoPolitic, PSG}.
A detailed overview of our dataset and our approach is depicted in Figure~\ref{fig:dataset_configuration}.

Our main contributions are as follows:

\begin{itemize}
\item We introduce a novel large-scale, multi-task Korean text classification dataset for improvements in the research fields of text classification for hatred and biased speech.
\item Our proposed architecture effectively solves multiple tasks simultaneously, reporting improved classification performance in each classification task.
\item We publicly provide all the resources, including the dataset and source code for academic research and real-world applications.
\end{itemize}

\section{Background and Related Work}

\subsection{Deep-Learning for Text Classification}
Traditional machine learning algorithms like Naïve Bayes~\cite{bayesian} and SVM~\cite{SVM_classification} have initially dominated text classification, relying on handcrafted features like bag-of-words.
While deep learning architectures such as CNNs and LSTMs have emerged and show improved performance~\cite{lstm, cnn_classification, multitask_recurrent, cnn_lstm_research}, they had limitations in capturing complex and long contexts.
These limitations could be overcome by the Transformer architecture~\cite{transformer}.
Some related studies demonstrate that pre-training a language model with large-scale datasets~\cite{xlnet, T5}, significantly improves Transformer-based model performance.
Additionally, adapted architectures like BigBird~\cite{bigbird} and LongFormer~\cite{longformer} excel in understanding sequences with over 4,096 tokens.
In this work, we utilize Korean pre-trained BERT-based models, including KR-BERT, KoBigBird, KoELECTRA, and RoBERTa (Korean ver.), to obtain state-of-the-art classification performance in Korean NLP tasks~\cite{krbert, kobigbird, park2020koelectra}.

\subsection{Hate Speech Classification in the Korean Language}

Text classification is a crucial tool in the NLP field.
Earlier work like Binary Relevance tackled classification but had scalability issues~\cite{binary_relevance}. Transformer-based models like BERT~\cite{bert} have since advanced the field, excelling in multi-label tasks. 
Among the classification tasks, hate speech detection is especially useful in online services.
However, classifying Korean hate speech is challenging because the Korean text contains unique linguistic features.
To solve this problem, some specialized datasets have been proposed.
For example, datasets like one with 9.4K entertainment news comments~\cite{beep} and another with 35,000 comments across multiple categories~\cite{kohs_dataset} have shown the effectiveness of BERT-based models. K-MHaS~\cite{K-MHaS}, with its 109k multi-labeled comments, also achieved optimal results with KR-BERT.
We note that our proposed dataset provides three unique tasks and adopts a more sophisticated annotation strategy.
Unlike existing works that mostly rely on binary classification, our dataset also provides a five-class ordinal classification for user preferences from \textit{hate} to \textit{love}.

\section{Proposed Dataset}
Our "KoMultiText" dataset is comprised of 150,000 Korean comments, more than 40,000 of which have been manually labeled by four human labelers following specific annotation guidelines. 
The remaining 110,000 comments are unlabeled.
The labeled comments are divided into a training dataset that contains about 38,000 comments and a test dataset that has 2,000 comments.
The labeled dataset encompasses a wide range of sentiment and biased comments, broadly categorized into three labels: \textbf{Preference}, \textbf{Profanity}, and \textbf{Specific Bias}.
For preprocessing, we applied minimal filtering to sentences that consisted solely of non-Korean languages, special characters, or emojis to preserve the original complexity of the dataset.

\subsection{Data Collection Pipeline}

To construct the dataset, we have sourced the comments from a forum, "Real-time Best Gallery", of DC Inside, a well-known online community in South Korea, utilizing web scraping techniques.
This forum has been selected due to its diversity and abundance of comments with various sentiments on a wide range of social topics.
We have collected 150,000 comments in a sequential manner without any selective curation. This approach ensures the dataset maintains a natural state of online discourse.

\subsection{Data Labeling Pipeline}

A team of four annotators has conducted the data labeling procedures.
During the labeling process, if any annotator encountered ambiguous or confusing comments, these were set aside for collective discussion.
In cases where there was disagreement among four annotators even after the discussion, a designated moderator made a final decision to ensure consistency.
With this method, we aim to achieve a high level of reliability and validity in our labeled dataset.

The criteria and details for each label are as follows.
\textbf{Preference (Multi-class)}:\hspace{0.5ex} Comments are labeled from 0 to 4 for representing sentiments from "Hate" to "Love". Each comment gets a single label, solely reflecting the writer's sentiment;
\textbf{Profanity (Binary-Class)}:\hspace{0.5ex} Comments are labeled 0 for "Without profanity" or 1 for "With profanity", which includes both traditional and newly-emerging offensive terms;
\textbf{Bias (Multi-label)}:\hspace{0.5ex} The 9 different types of biases are labeled as individual categories in each comment.
A binary value of 0 for "Without Bias" and 1 for "With Bias" is assigned for each bias type.
The detailed information of each bias type is described in the supplementary material due to the space limitation of the paper.

\subsection{Data Distribution}

\begin{figure*}[htp]
    \centering
    \centerline{\includegraphics[width=1\textwidth]
    {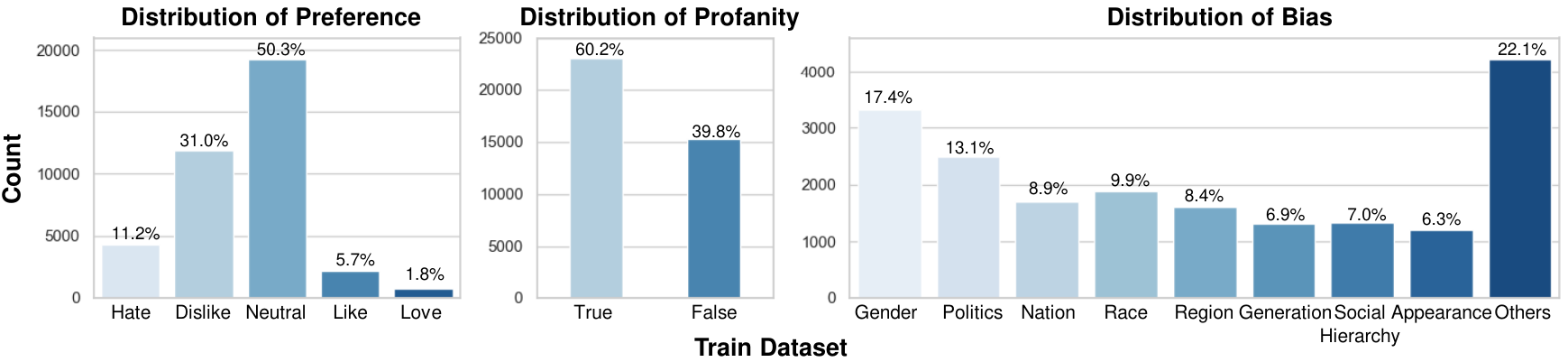}}
    \caption{Distribution graph of train dataset. Each text contains three kinds of annotations: (1) Preference label (multi-class). (2) Bias labels (multi-label). (3) profanities (binary class).}
    \label{fig:train_data_distribution}
\end{figure*}

The total labeled dataset consists of more than 40,000 comments and shows a significant imbalance across various classes.
This imbalance is attributed to the non-selective collection of the data. Figure~\ref{fig:train_data_distribution} displays the label distribution in the 38,000 comments designated for the training dataset.
Specifically, "Like" and "Love" within the \textbf{Preference} labels are highly under-represented.
Likewise, certain biases have insufficient representation in the distribution of \textbf{Bias} labels.
Despite these issues, we have attempted to balance the label distribution in the test dataset.
Further details, including a figure showing a more balanced distribution in the test dataset, are provided in the supplementary material.
Although matching the label distribution of the test dataset with the training dataset could improve performance metrics, our study prioritizes providing a reliable and robust evaluation benchmark.

\section{Experiments}

\subsection{Models}

\begin{figure*}[htp]
    \centering   \centerline{\includegraphics[width=1.25\textwidth]{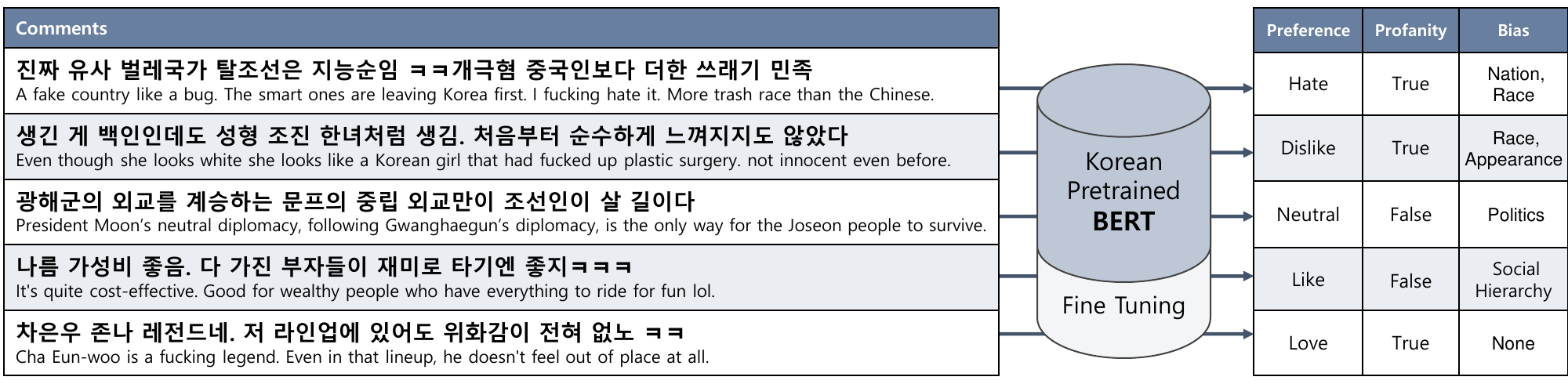}}
    \caption{The illustration of our proposed model with the input and output examples.}
    \label{fig:model}
\end{figure*}

We design multi-task models to simultaneously address three tasks (Preference, Profanity, and Bias) in a single training session.
Figure ~\ref{fig:model} illustrates an example of the model's input comments and the corresponding label outputs for the three heads: \textbf{Preference}, \textbf{Profanity}, and \textbf{Bias}.
We utilize four different Korean pre-trained BERT-based models for our experiments: KR-BERT, KoBigBird, KoELECTRA, and RoBERTa (Korean ver).
These models are sourced from the Hugging Face model hub, which is a broadly adopted repository in NLP research fields.
We conduct tokenization using the respective tokenizers provided by Hugging Face.

To further investigate the capabilities of our models, we have conducted experiments in both single-task and multi-task settings.
In the single-task setting, each model is trained to focus solely on one of the three tasks: \textbf{Preference}, \textbf{Profanity}, or \textbf{Bias}.
In contrast, the multi-task setting involves training the model to simultaneously learn across all three tasks using separate heads for \textbf{Preference}, \textbf{Profanity}, and \textbf{Bias}.
By adopting these two different training paradigms, we aim to gain a comprehensive understanding of each model's performance capabilities.

We have experimented with \(3 \times 10^{-6}\) initial learning rate, and dynamically adjusted it using a linear rate scheduler with a 10\% warm-up period.
We adopted the AdamW optimizer~\cite{AdamW}, applying weight decay to all parameters except for bias and LayerNorm weights. 
Additionally, to address the data imbalance issue in the \textbf{Preference} and \textbf{Bias} tasks, we apply different loss weights per category to emphasize underrepresented classes. Moreover, we employ class weights within the loss functions, which ensures that the model allocates more significance to the minority classes. These approaches contribute to achieving a more balanced performance across various categories.

\subsection{Evaluation Metrics}

For the task of \textbf{Preference}, which is a multi-class problem with five classes, we employ Accuracy and F1-score as the primary metrics for evaluation~\cite{metrics}.
In addition to these conventional metrics, we also report Top-2 Accuracy and Mean Absolute Error (MAE) to account for the ordinal nature of the Preference labels.
Top-2 Accuracy refers to the approach that considers a prediction as \textit{correct} if the true label is among the two classes with the highest confidence scores predicted by the model. 
These supplementary metrics enable a more nuanced understanding of the model's performance, ensuring that our evaluation is comprehensive and precise.

For the tasks of \textbf{Profanity} and \textbf{Bias}, which are essentially binary classification problems, we utilize the Area Under the Receiver Operating Characteristic Curve (AUROC) and F1-score as our evaluation metrics~\cite{AUROC, Enhancing_bi,  sentiment, analysis_classification}. 
Particularly for the Bias task, we additionally employ the Precision-Recall (PR) curve to enhance evaluation reliability in datasets with a limited number of positive instances~\cite{PRROC}.

\subsection{Results}

\begin{table*}[!ht]
    \centering
    \flushleft
    \begin{minipage}{0.5\textwidth}
    {\fontsize{11}{14}
    \caption{The overall classification performance for the Preference, Profanity, and Bias.}
    \renewcommand\arraystretch{1.5}  
    \begin{adjustbox}{width=10.0cm, center}
    \begin{tabular}{c|cccc|cc|cc} \toprule  
    \multirow{4}{*}{\textbf{Architectures}} & \multicolumn{7}{c}{\centering\textbf{Test Dataset Results}} \\
    \cline{2-9}  
    & \multicolumn{4}{c|}{\textbf{Preference}} & \multicolumn{2}{c|}{\textbf{Profanity}} & 
    \multicolumn{2}{c}{\textbf{Bias}} \\
    \cline{2-9}  
    \cline{2-9}  
    & \multicolumn{4}{c|}{\textbf{Multi-Class Classification}} &  \multicolumn{4}{c}{\textbf{Binary Classification}} \\  
    \cline{2-9}  
    &Accuracy & Top-2 Accuracy& \multicolumn{1}{c}{F1-macro} & MAE & AUROC & F1-score & AUROC & PRROC\\
    \cline{1-9}  
    \multirow{13}{*}{}
    RoBERTa (Single) & 75.74 &	92.57 &	75.28 &	0.33 &	97.96	&94.35 	&	92.65 &	74.88 \\  
    RoBERTa (Multi)&70.54&	90.10&	69.06&	0.35&	97.03&	92.70&	92.07&	73.92\\
    KoBigBird (Single) & 78.47 &	93.81 &	77.25 &	0.24 &	98.75 &	95.4	 &	94.46 &	78.17\\   		
    KoBigBird (Multi) & 69.80 &	90.09 &	67.90 &	0.39 &	97.71 &	94.90 &	93.15 & 75.24\\
    KoELECTRA (Single) & 76.24 & 93.56 & 75.17 & 0.27 &	98.75&	95.81&	94.28&	75.00\\
    KoELECTRA (Multi)  &65.35&	88.86&	63.51&	0.43&	94.59&	88.54&	93.27&	74.64\\
    KR-BERT (Single) & 74.01 & 94.55 & 72.35 & 0.27 & 98.21&	96.33  & 94.13 & 78.16\\ 
    KR-BERT (Multi) & 72.28 & 93.56 & 70.88 & 0.32 & 98.43 & 95.84 &	94.34 & 78.22 \\
    \bottomrule
    \end{tabular}
    \end{adjustbox} 
    \label{tab:preference_profanity_table}}
    \end{minipage}%
    \begin{minipage}{0.75\textwidth}
    \flushright
    \captionof{table}{\parbox{.4\textwidth}{Comparison of Single \\ and Multi-Task models.}}
    \renewcommand\arraystretch{1.4}
    \begin{adjustbox}{width=6.5cm, center}
    \centering
    \begin{tabular}{c|c|c|c|c}
    \hline
    \multirow{2}{*}{\textbf{Model}}& \multicolumn{2}{c|}{\textbf{Single-Task}} & \multicolumn{2}{c}{\textbf{Multi-Task}}\\
    \cline{2-5}
     & Total Params & Size (MB) & Total Params & Size (MB) \\
    \hline
    KoBigBird & 343,927,311 & 1,311.98 & 113,765,391 & 433.94 \\
    RoBERTa & 339,657,999 & 1,295.69 & 112,342,287 & 428.55 \\
    KoELECTRA & 334,520,079 & 1,276.10 & 110,629,647 & 422.02 \\
    KrBERT & 306,869,775 & 1,170.62 & 101,412,879 & 386.86 \\
    \hline
    \end{tabular}
    \end{adjustbox} 
    \vskip\baselineskip
    \label{tab:single_multi_comparison_table}
    \end{minipage}
\end{table*}

Table~\ref{tab:preference_profanity_table} presents the overall performance for the \textbf{Preference}, \textbf{Profanity}, and \textbf{Bias} tasks across four different models, comparing both single-task and multi-task settings.
In most experiments, the single-task setting outperforms the multi-task setting.
On the other hand, Table~\ref{tab:single_multi_comparison_table} illustrates the computational benefits of multi-task settings.

\textbf{Computational Advantages of Multi-Task Models}\hspace{0.5ex} 
Despite generally lower classification performance, multi-task models show substantial efficiency in computation compared to single-task models.
For instance, the multi-task of all three models requires approximately 33\% of the total parameters and the memory size compared to single-task.
This computational efficiency not only leads to a significant reduction in resource requirements but also accelerates training and inference times.

\textbf{Evaluation for the Preference Task}\hspace{0.5ex} 
For the Preference Task, single-task models achieve the best accuracy of 78.47\% and an F1-score of 77.25, outperforming the multi-task models, which show 71.78\% and 70.58.
Additional metrics like Top-2 Accuracy and MAE provide additional evaluation information.
In the single-task, the models achieve the best Top-2 Accuracy of 94.55\% and an MAE of 0.24, compared to 93.56\% and 0.32 in the multi-task.
Considering the ordinal nature of the task and that a one-point difference in preference is feasible in human judgment, these results are highly encouraging.

\textbf{Evaluation for the Profanity Task}\hspace{0.5ex} 
For the Profanity Task, both single-task and multi-task models demonstrate the highest performance across all tasks.
The highest AUROC and F1-scores are 95.58 and 95.75 for single-task settings, and 94.29 and 94.64 for multi-task settings.
This superior performance is likely due to consistent linguistic patterns in Profanity. While this may resemble a rule-based approach, our models also excel in recognizing new slang, typos, and contextual nuances.

\textbf{Evaluation for the Bias Task}\hspace{0.5ex} 
The models demonstrate varying performance and trade-offs across different biases.
The average AUROC and PRROC scores for all biases are presented in Table~\ref{tab:preference_profanity_table}. Despite the presence of imbalances in Bias categories, the models exhibit significant classification performance improvement. 
The 'Others' category, in particular, poses the most substantial training challenge due to the diversity of bias topics it encompasses. 
However, the results suggest that acquiring sufficient data could further enhance the models' ability to detect specific biases.
Detailed AUROC, PRROC, and F1-scores for each bias type are available in the supplementary material.

\section{Conclusion}

We introduce "KoMultiText", a comprehensive, multi-task Korean text dataset, designed to enhance online moderation by detecting biased and hateful speech. By employing advanced transformer models, we achieve superior performance compared to human readers across various classification tasks. Despite encountering challenges such as class imbalance and annotation bias, our work represents a significant advancement in the field of text classification, particularly for Korean content. Through the public release of our dataset and models, we aim to encourage the development of real-world applications that can improve online discourse and foster community well-being. Our work not only advances the field of Korean text classification but also establishes a benchmark for the socially responsible application of language models.

\bibliographystyle{plain}
\bibliography{neurips_2023}

\newpage

\appendix

\section{Dataset Anlaysis}

\textbf{Dataset Description and Future Directions}

\textit{DC Inside} is a popular online community in South Korea, founded in 1999. This web service is one of the oldest and largest internet forums in South Korea, covering a wide range of topics from politics, news, and technology to entertainment and hobbies. The forum has various sub-forums, among which the "Real-time Best Gallery" has been known for its active discussions and diverse opinions. This forum not only allows us to capture a broad spectrum of sentiments but also provides a snapshot of prevalent issues in contemporary South Korean online culture. DC Inside has its own terms of service, which we have reviewed to ensure compliance. Our web scraping technique has been designed thoughtfully to be minimally invasive and respectful of the website's terms and user policies. As of September 2023, DC Inside hosts 62,681 individual galleries, with over 800,000 new posts and more than 2,000,000 new comments added daily. For more detailed information, the website can be accessed at \url{https://www.dcinside.com/}.

While our current dataset provides valuable insights into the state of online discourse in a specific South Korean online community, we recognize the need for more diversified data to generalize our work. Future studies will aim to include comments from additional platforms such as YouTube, online news, and other SNS platforms. Through our further expansion, we expect to create a dataset more representative of broader online discourse in South Korea for future research. \\

\textbf{Data Distribution of Test Dataset}

Compared to the highly imbalanced training dataset, the test dataset has been constructed to follow as uniform a distribution as possible. Due to the inherent characteristics of the data and the multi-tasking setting, achieving a perfectly balanced dataset is not feasible. We note that our test dataset represents a great approximation under realistic conditions. To enhance the reliability of the evaluation, the labelers collaboratively reviewed and revised the annotations of the test dataset.

\vspace{0.5cm}
\begin{figure*}[htp]
    \centering
    \centerline{
    \includegraphics[width=1.0\textwidth]{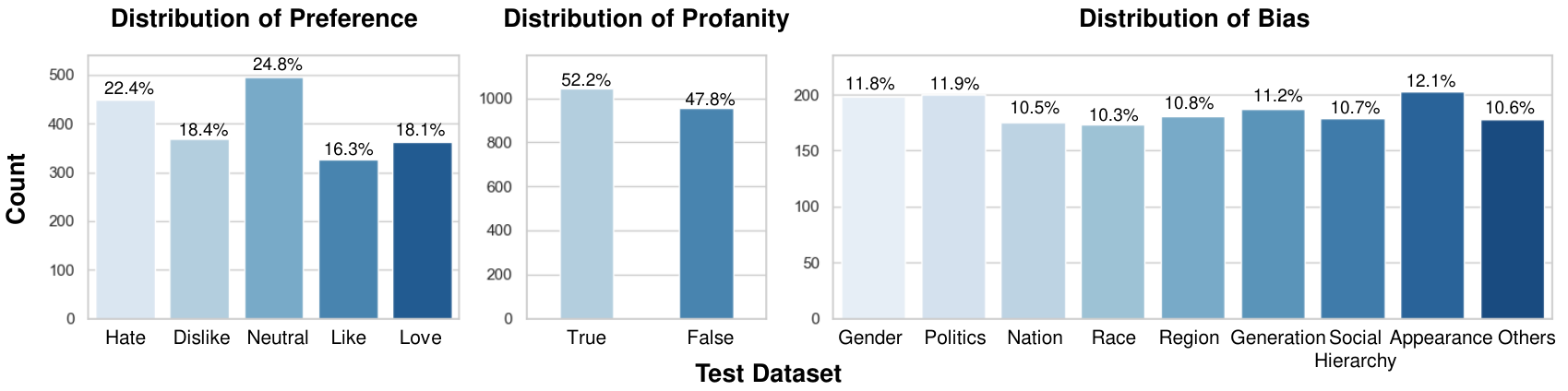}}
    \caption{Distribution graph of the test dataset. Each text contains three kinds of annotations: (1) Preference label (multi-class). (2) Bias labels (multi-label). (3) profanities (binary-class).}
    \label{fig:test_distribution}
\end{figure*}

\newpage

\begin{table*}[!ht]
\vskip\baselineskip

\caption{Data distribution of \textit{Bias} annotations in training and test dataset. The training dataset consists of 38K comments and the test dataset consists of 2K comments. The test dataset contains 1,674 bias annotations and the training dataset contains 19,063 Bias annotations. Due to the multi-task learning property, some text samples only contain \textit{Preference} or \textit{Profanity} in our proposed dataset. Therefore, the total number of \textit{Bias} annotations could be smaller than the total number of text samples.}

\vskip\baselineskip
\renewcommand\arraystretch{1.2}  
\begin{adjustbox}{width=10.0cm, center}
\begin{tabular}{l|rr} 
\hline
\multirow{1}{*}{\textbf{Class}} & {\textbf{Count - Training (\%)}} & {\textbf{Count - Test (\%)}} \\ \hline
Gender & 3315 (17.4\%) & 198 (11.8\%) \\
Politics & 2492 (13.1\%) & 200 (11.9\%) \\
Nation & 1697 (8.9\%) & 175 (10.5\%) \\
Race & 1881 (9.9\%) & 173 (10.3\%) \\
Region & 1610 (8.4\%) & 181 (10.8\%) \\
Generation & 1313 (6.9\%) & 187 (11.2\%) \\
Social Hierarchy & 1333 (7.0\%) & 179 (10.7\%) \\
Appearance & 1205 (6.3\%) & 203 (12.1\%) \\
Others & 4217 (22.1\%) & 178 (10.6\%) \\
\hline
\end{tabular}
\end{adjustbox}
\label{tab:dataset_distributiuon_table}
\vskip\baselineskip
\end{table*} 

\vspace{2.0cm}

\begin{figure*}[htp]
    \caption{The frequencies of the top-8 keywords for each bias label.}
    \centering
    \centerline{
    \includegraphics[width=1.3\textwidth]{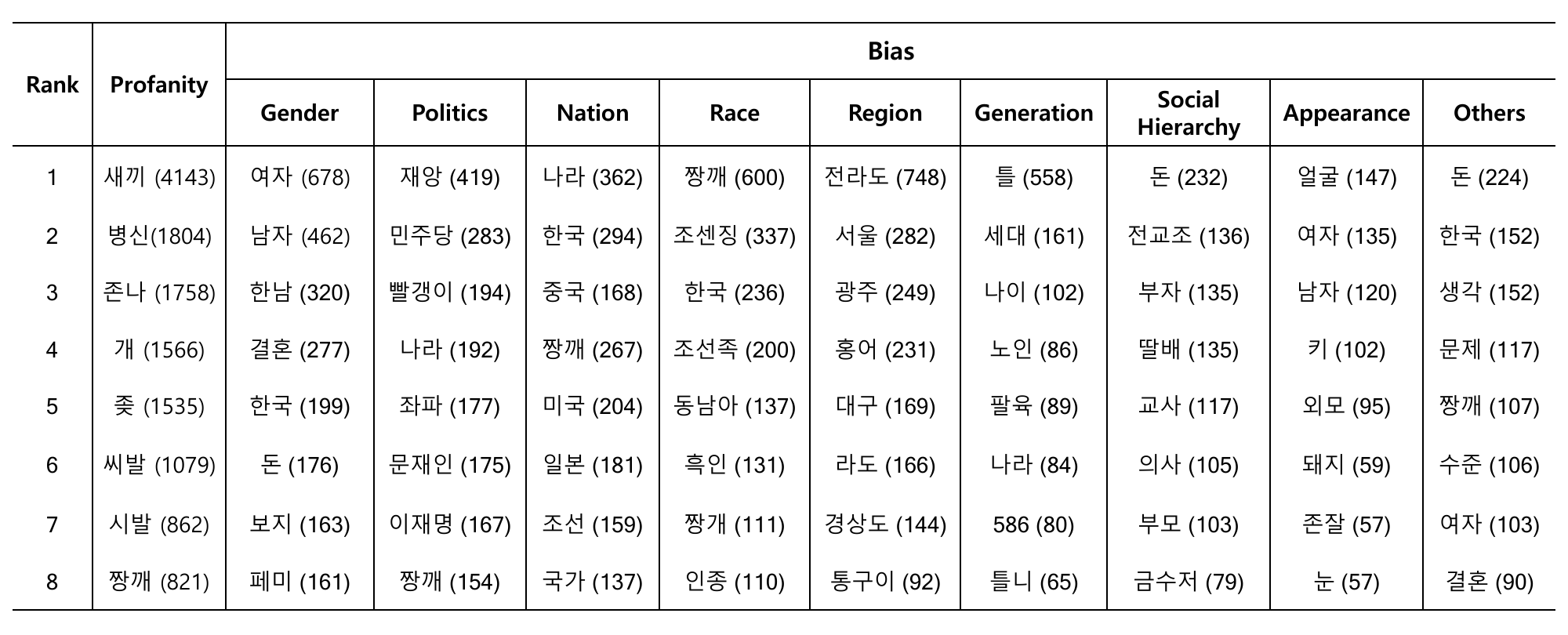}}
    \label{fig:keyword8}
\end{figure*}

\newpage
\begin{table}[h]
\caption{Description of Bias Labels}
\vskip\baselineskip
\begin{adjustbox}{width=1.1\textwidth,center}
\renewcommand{\arraystretch}{1.4} 
\centering
\begin{tabular}{|c|m{12cm}|}
\hline
\textbf{Bias Type} & \textbf{Description} \\
\hline
Gender Bias & Comments that show prejudice or differential treatment based on someone's gender or sexual orientation, including derogatory terms or phrases. \\
\hline
Political Bias & Comments that display prejudice against or favoritism towards affiliations or individuals based on political ideologies. \\
\hline
National Bias & Comments that favor or discriminate based on nationality, including stereotypes related to nationality or derogatory portrayals. \\
\hline
Racial Bias & Comments that exhibit prejudice based on race or ethnic background, including racial slurs or stereotypical portrayals. \\
\hline
Regional Bias & Comments showing prejudice towards individuals or groups based on a specific geographic region they come from within a country, such as local stereotypes. \\
\hline
Generational Bias & Comments showing prejudice based on age group or generational cohort, like generalizations or stereotypes of younger or older people. \\
\hline
Social Hierarchy Bias & Comments that discriminate based on someone's social or economic status, such as job, income level, or educational background. \\
\hline
Appearance Bias & Comments that show bias based on physical appearance, such as attractiveness, clothing, body size, or other physical characteristics. \\
\hline
Other Biases & Comments that exhibit other types of biases, including but not limited to religion, occupation, animals, and specific communities.
\\
\hline
\end{tabular}
\end{adjustbox}
\vskip\baselineskip
\end{table}

\section{Analysis on the Experimental Results}

We provide a detailed analysis of the model's performance across various biases, as well as a comprehensive visualization of the KR-BERT's multi-task training process based on the test dataset.

\subsection{Biases Classification Performance}

Table~\ref{tab:bias_table} provides a comprehensive breakdown of the model's performance for each bias category. This table includes AUROC (Area Under the Receiver Operating Characteristic Curve), F1 score, and PRROC (Precision-Recall curve) values for each bias. These metrics provide additional information that is useful to interpret the experimental results and help the understanding of how the model performs for each specific type of bias, which is essential given the diversity and intricacies of the biases present in the dataset.
Different model architectures show varying performance across bias labels. While single-task models generally achieve superior performance on average for specific biases, the multi-task models also demonstrate competitive classification performance and tend to exhibit more consistent results across the diverse range of biases. This result suggests that, although single-task models reach peak performance in certain tasks, multi-task models offer a more reliable performance across a broader spectrum of tasks because the multi-task learning method leverages more abundant annotation information per text, which can result in improved feature representation learning.

Specific biases, such as 'Politics' and 'Nation', consistently achieve high scores across architectures, while categories like 'Appearance' and 'Others' present challenges to obtaining improved classification performance. We note that the diminished performance in the 'Appearance' category could be primarily due to limited training data. Moreover, the 'Others' category also faces challenges from including diverse biases and insufficient data for certain subcategories. Defining all the possible other biases is fundamentally hard. Gathering more data for these categories could significantly enhance the model's classification performance and robustness.

\begin{sidewaystable}
\caption{Detailed AUROC, F1-score, and PRROC results for each specific bias type.}
\vspace*{0.2cm}
\renewcommand\arraystretch{2.3}  
{\Huge
\begin{adjustbox}{width=22.0cm, center}
\begin{tabular}{c|ccc|ccc|ccc|ccc|ccc|ccc|ccc|ccc|ccc} \toprule
\multirow{4}{*}{\textbf{Architecture}} & \multicolumn{27}{c}{\textbf{Test Dataset Results}} \\ \cline{2-28} & \multicolumn{27}{c}{\textbf{Multi-label Classification}} \\ \cline{2-28}
& \multicolumn{3}{c|}{\textbf{Gender}} & \multicolumn{3}{c|}{\textbf{Politics}} & \multicolumn{3}{c|}{\textbf{Nation}} & \multicolumn{3}{c|}{\textbf{Race}} & \multicolumn{3}{c|}{\textbf{Region}} & \multicolumn{3}{c|}{\textbf{Generation}} & \multicolumn{3}{c|}{\textbf{Social Hierachy}} & \multicolumn{3}{c|}{\textbf{Appearance}} & \multicolumn{3}{c}{\textbf{Others}} \\ \cline{2-28} 
& AUROC & F1-score & PRROC & AUROC & F1-score & PRROC & AUROC & F1-score & PRROC & AUROC & F1-score & PRROC & AUROC & F1-score & PRROC & AUROC & F1-score & PRROC & AUROC & F1-score & PRROC & AUROC & F1-score & PRROC & AUROC & F1-score & PRROC\\
\midrule
RoBERTa (single)  &93.55&	73.04&	77.9&	97.78&	88.00&	89.16&	96.52&	69.47&	69.77&	97.67&	76.71&	78.92&	98.63&	86.96&	94.48&	94.22&	84.00&	86.48&	89.84&	63.01&	68.42&	86.46&	68.75&	65.39&	79.16&	45.05&	43.38 \\
RoBERTa (multi) &94.02&	70.80&	72.94&	96.43&	84.21&	89.06&	94.49&	65.75&	68.23&	95.91&	78.87&	82.34&	95.02&	83.72&	87.93&	94.68	&83.67&	86.94&	89&	66.67&	71.63&	89.77&	71.43&	68.48&	79.31&	44.9&	37.75\\
KoBigBird (single) &93.16&	74.29&	78.71&	98.42&	88.17&	92.53&	97.18&	77.11	&87.94&	98.42&	79.41&	75.51&	93.73&	75.56&	82.55&	97.81	&84.44&	91.67&	92.54&	70.27&	74.05&	93.06&	72.34&	72.41&	85.86&	48.89&	48.13\\
KoBigBird (multi) &94.47&	76.11&	81.57&	96.64&	79.61&	88.03&	95.65&	77.78	&75.69&	98.48&	73.68&	78.15&	98.21&	85.11&	91.68&	95.56	&81.72&	88.76&	89.16&	66.67&	69.69&	88.10&	61.86&	56.62&	82.05&	40.94&	47.00\\
KoELECTRA (single)&93.71&	70.49&	73.19&	98.62&	83.02&	93.4&	97.23&	72.53	&80.48&	98.38&	65.12&	81.29&	99.06&	83.33&	89.97&	96.82	&81.55&	89.92&	91.32&	61.18&	66.33&	87.95&	70.27&	62.13&	85.43&	36.36&	38.27\\
KoELECTRA (multi) &95.81&	74.34&	80.16&	97.75&	80&	90.59&	97.36&	75.56&	78.82&	97.99&	68.35&	82.25&	98.94&	80&	90.86&	93.84&	82.83&	85.12&	87.53&	60.53&	66.08&	87.55&	71.15&	64.23&	82.67&	37.31&	33.65\\
KR-BERT (single) &93.16 & 74.29&	78.71&	98.42&	88.17&	92.53&	97.18&	77.11&	87.94&	98.42&	79.41&	75.51&	93.73&	79.41&	82.55&	97.81& 84.44&	91.67&	92.54&	70.27&	74.05&	93.06&	72.34&	72.41&	85.86 &	48.89&	48.13\\
KR-BERT (multi) &93.27	&77.36&	79.50&	98.13&	85.71&	91.96	&97.36 &82.35&	88.68&	98.81&	77.14&	77.56&	92.05&	76.19&	79.39	&98.03	&85.39&	92.03&	91.93&	68.66&	73.27&	93.89&	73.47&	74.73&	85.56&	49.46&	46.89\\
\bottomrule
\end{tabular}
\end{adjustbox}}
\label{tab:bias_table}
\end{sidewaystable}

\subsection{Training Progress of KR-BERT in the Multi-task Setting}
The training dynamics of models in multi-task settings provide essential information about a model's overall performance, optimization process, and potential issues. Analyzing the learning curves allows us to assess the stability of the training, detect possible overfitting, and determine the model's learning rate.
In this subsection, we focus on the training progress of the KR-BERT model due to its standout average performance in the experiments. The subsequent figures detail the multi-task learning curves of KR-BERT, covering key metrics such as training loss, task-specific accuracy, F1 score, AUROC, PRROC, and others. We note that all these results are derived from the test dataset.

\begin{minipage}{\textwidth}
    \centering
    \centerline{
    \includegraphics[width=1.1\linewidth, trim=1.5cm 0 0 0]{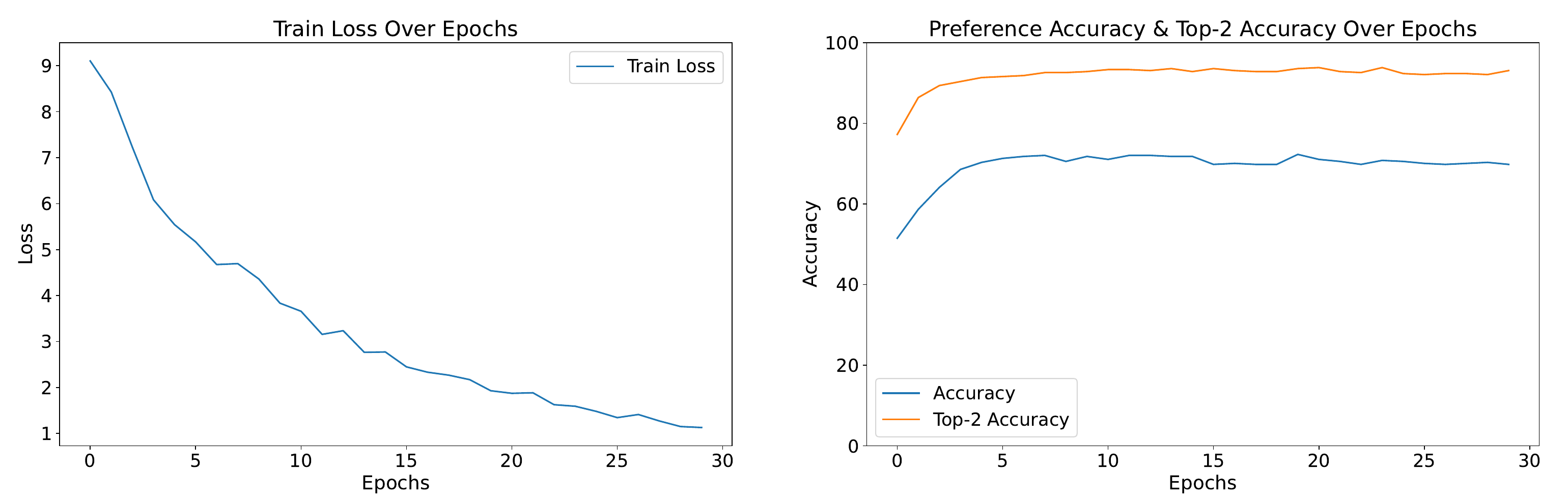}}
    \begin{adjustbox}{width=1.0\linewidth, center}
    \begin{minipage}{0.44\linewidth}
        \centering
        \vspace*{0.1cm}
        \captionof{figure}{The training loss curve of our multi-task model.}
        \label{fig:fig6_left}
    \end{minipage}\hspace{2.2cm}\hfill
    \begin{minipage}{0.45\linewidth}
        \centering
        \vspace*{0.1cm}
        \captionof{figure}{The accuracy and Top-2 accuracy curves for the \textit{Preference Task.}}
        \label{fig:fig7_right}
    \end{minipage}
    \end{adjustbox}
\end{minipage}

Figure~\ref{fig:fig6_left} illustrates the overall training loss dynamics of the multi-task model. Throughout the 30 epochs, the consistent decrease in loss indicates that the model learns effectively. However, despite the diminishing loss, extending the training beyond these epochs does not lead to improvements in other performance metrics. We note that the convergence of the training loss sometimes does not directly interpret the classification performance improvement due to the over-fitting issue.

Figure~\ref{fig:fig7_right} provides the Top-1 accuracy and Top-2 accuracy of the trained model for the \textit{Preference} Task. After the 5th epoch, both the Accuracy and Top-2 Accuracy remain stable without significant fluctuations. The Top-2 Accuracy consistently records about 20\% higher than the Accuracy. The maximum Accuracy achieved is 72.28, while the Top-2 Accuracy peaks at 93.56.

\begin{minipage}{\textwidth}
    \centering
    \centerline{
    \includegraphics[width=1.1\linewidth, trim=2.0cm 0 0 0]{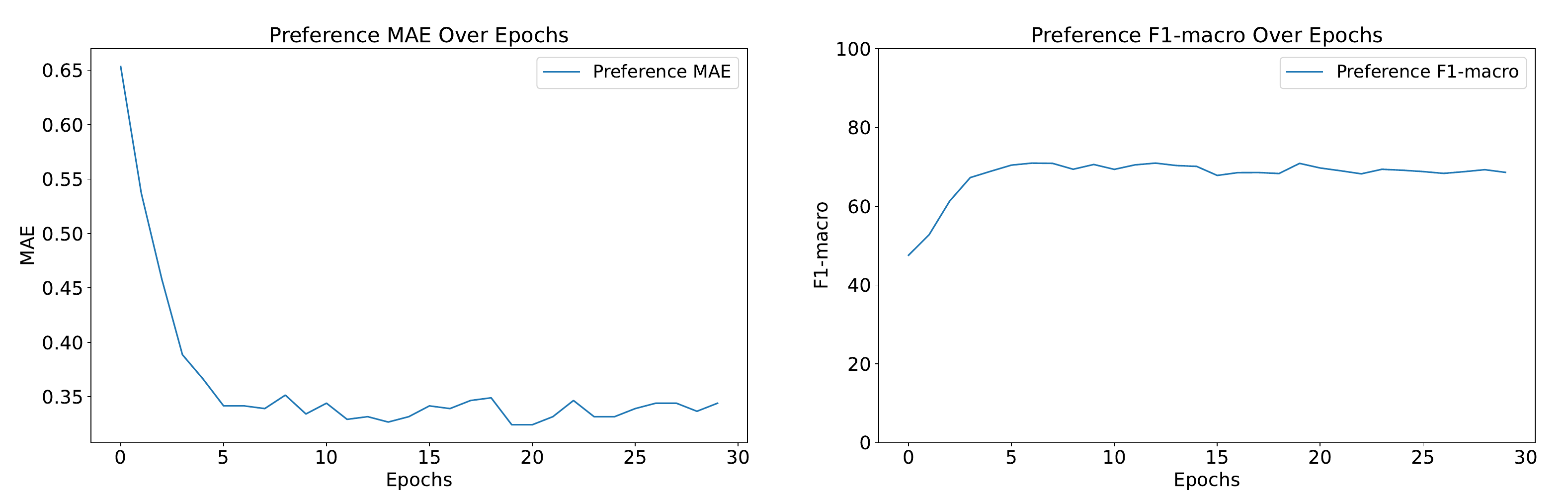}}
    \begin{adjustbox}{width=1.00\linewidth, center}
        \begin{minipage}{0.43\linewidth}
            \centering
            \vspace*{0.1cm}
            \captionof{figure}{The MAE of the model for the \textit{Preference} classification task.}
            \label{fig:fig8_left}
        \end{minipage}\hspace{2.5cm}\hfill
        \begin{minipage}{0.43\linewidth}
            \centering
            \vspace*{0.1cm}
            \captionof{figure}{The F1-macro scores of the model for the \textit{Preference} classification task.}
            \label{fig:fig9_right}
        \end{minipage}
    \end{adjustbox}
\end{minipage}

Figure~\ref{fig:fig8_left} illustrates the Mean Absolute Error (MAE) of the model for the \textit{Preference} task. This result records a minimum value of 0.32. Considering that one class difference in the \textit{Preference} Task corresponds to a numeric difference of 1, this result indicates a notably low value, signifying the model's efficiency in capturing subtle variations in preferences.

Figure~\ref{fig:fig9_right} depicts the F1-macro performance of the \textit{Preference} task. Consistent with other learning curves for the \textit{Preference} task, this result demonstrates stability without significant fluctuations after the 5th epoch. The highest recorded F1-macro value is 70.58.

\begin{figure*}[htp]
\captionsetup{margin={1cm, 0cm}}
    \begin{adjustwidth}{-4.8cm}{-3cm} 
        \centering
        \begin{minipage}[b]{0.49\textwidth}
            \centering
            \includegraphics[width=1.15\linewidth]{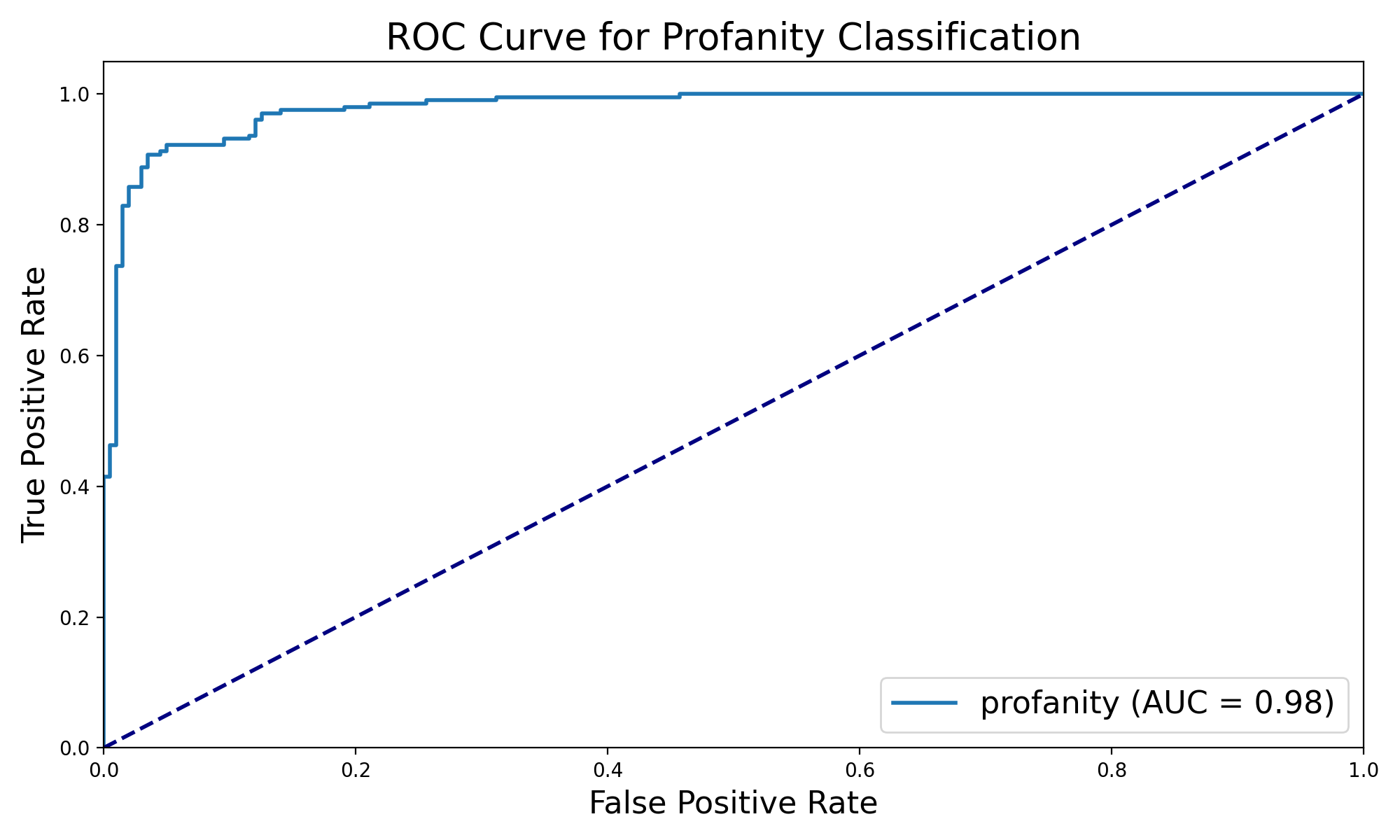}
            \caption{The ROC curve for the \textit{Profanity} task.}
            \label{fig:fig10_left}
        \end{minipage}
        \hspace{0.06\textwidth}
        \begin{minipage}[b]{0.49\textwidth}
            \centering
            \includegraphics[width=1.18\linewidth]{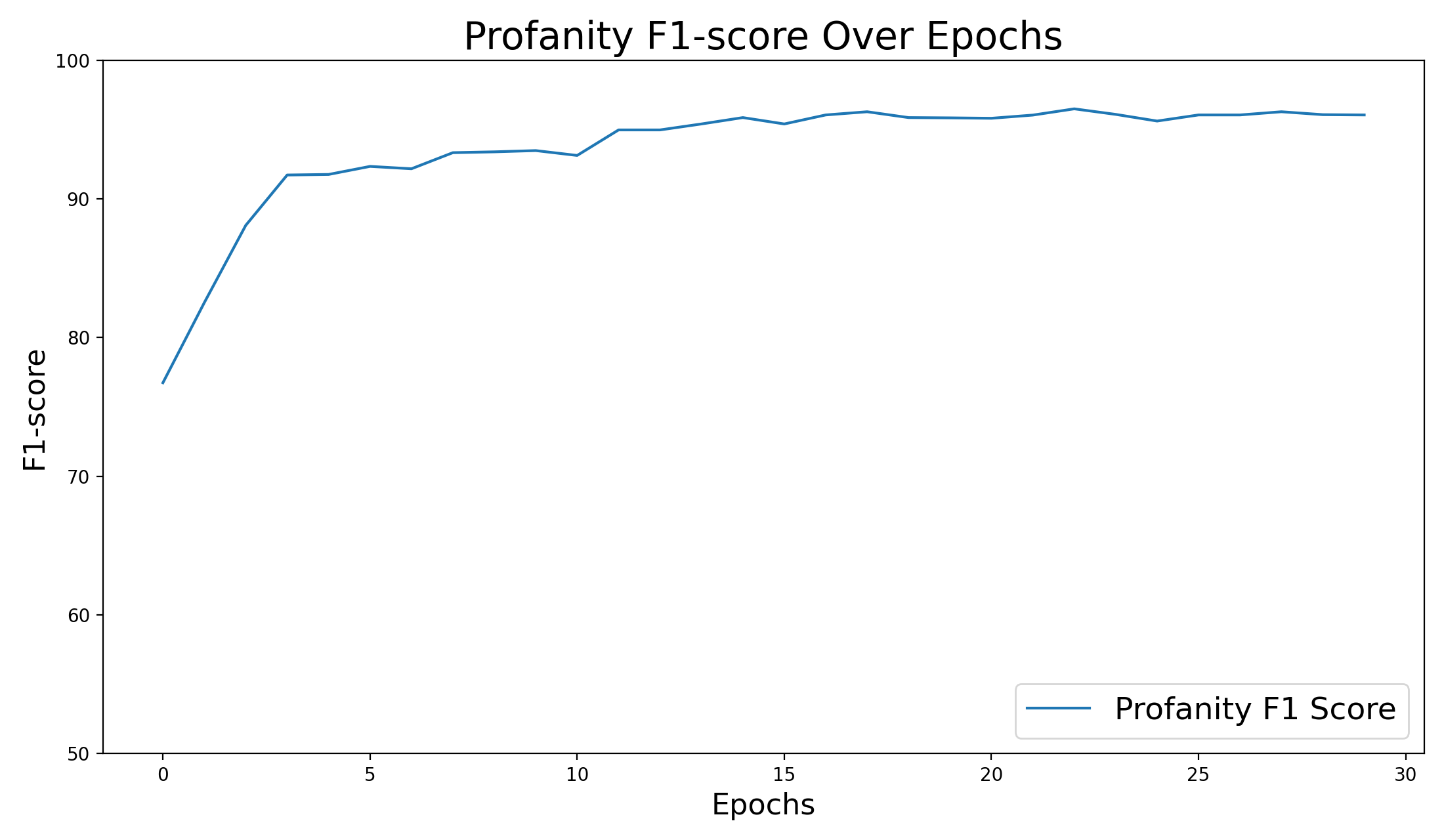}
            \caption{The F1-scores of the model for the \textit{Profanity} task.} 
            \label{fig:fig11_right}
        \end{minipage}
    \end{adjustwidth}
\end{figure*}

\begin{figure*}[htp]
\captionsetup{margin={1cm, 0cm}}
    \begin{adjustwidth}{-5.0cm}{-3cm} 
        \centering
        \begin{minipage}[b]{0.48\textwidth}
            \centering
            \includegraphics[width=1.20\linewidth]{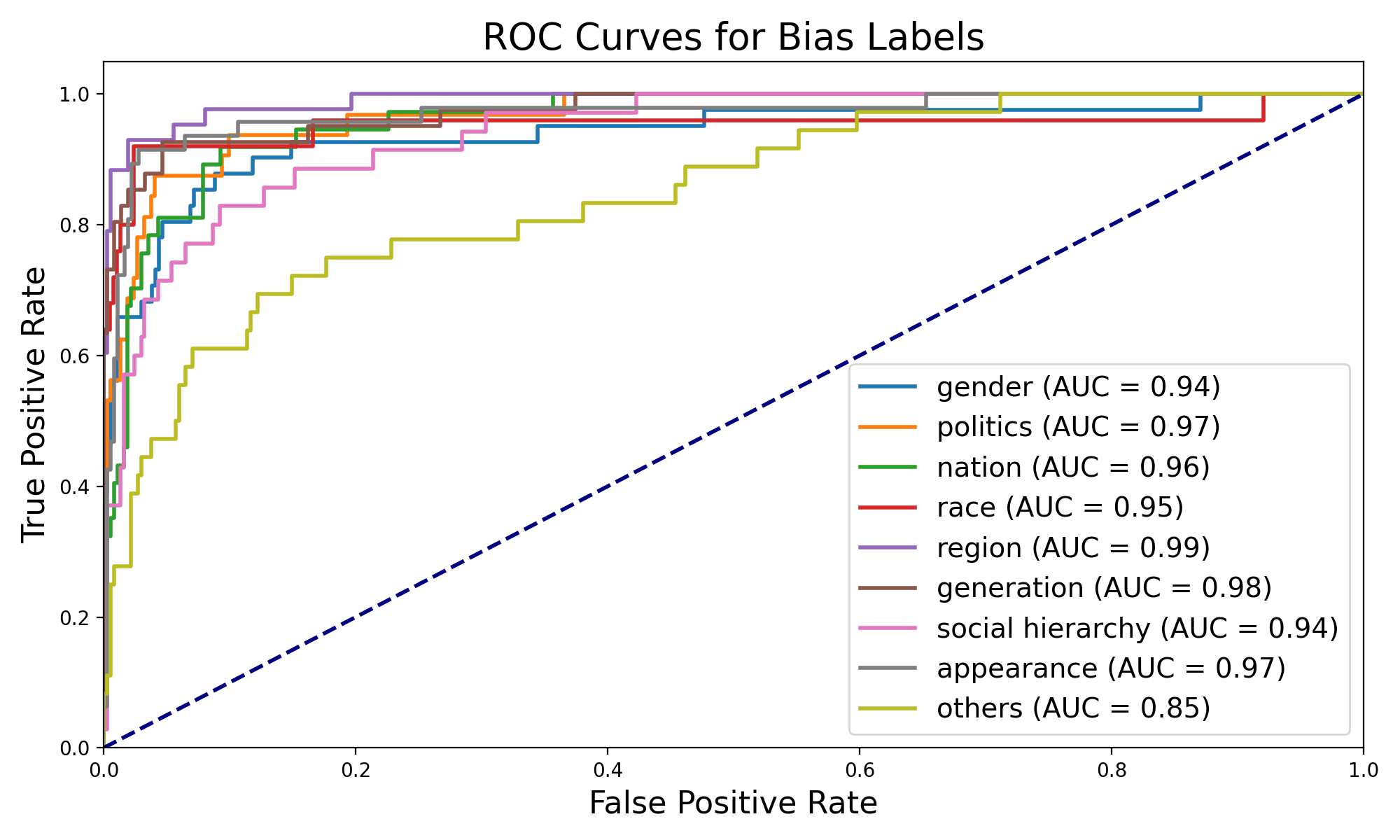}
            \caption{The ROC curve for the overall \textit{Bias} labels.}
            \label{fig:fig12_left}
        \end{minipage}
        \hspace{0.08\textwidth}
        \begin{minipage}[b]{0.48\textwidth}
            \centering
            \includegraphics[width=1.20\linewidth]{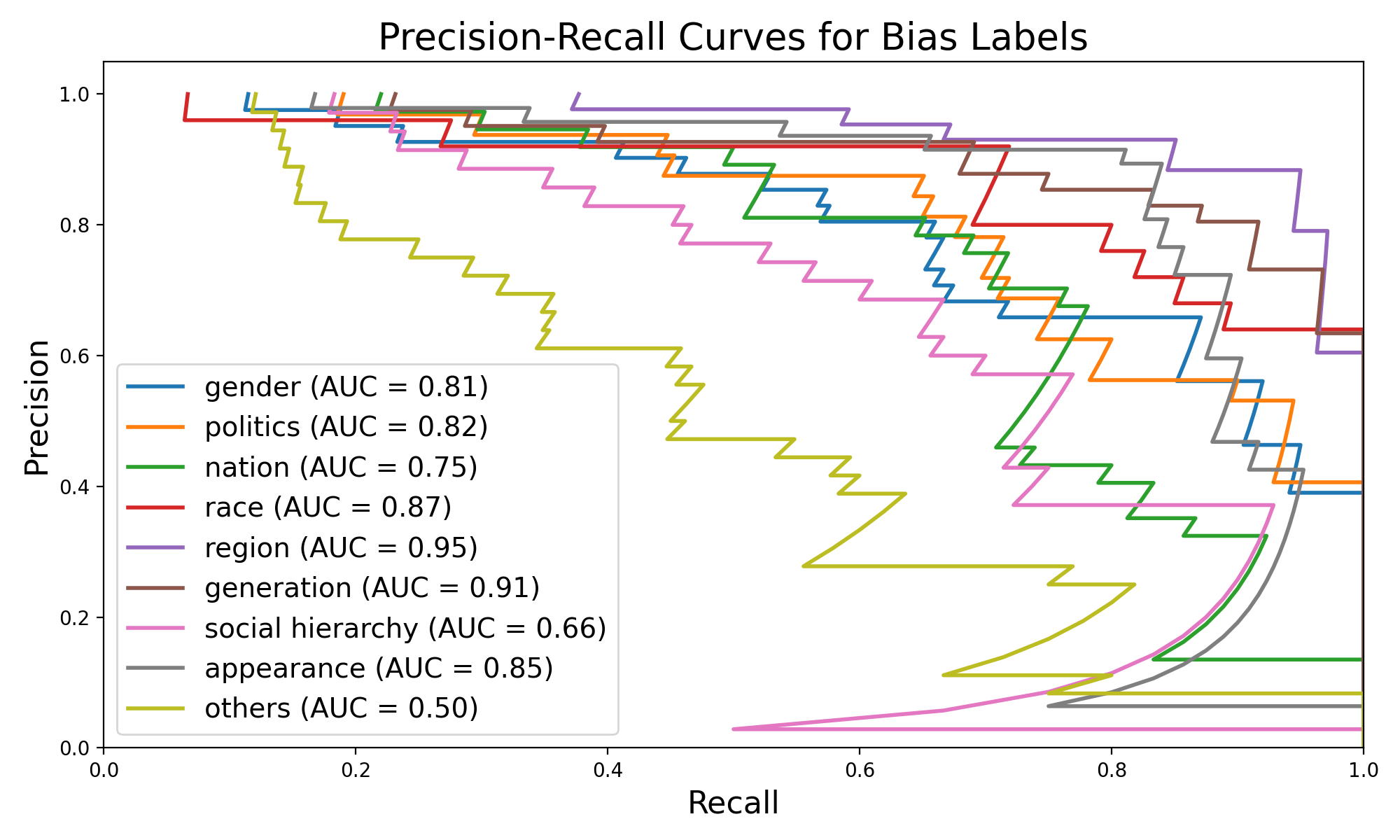}
            \caption{The PRROC for the overall \textit{Bias} labels.}
            \label{fig:fig13_right}
        \end{minipage}
    \end{adjustwidth}
\end{figure*}

Figure~\ref{fig:fig10_left} shows the ROC curve for the \textit{Profanity} task. The AUROC curve is a comprehensive metric that evaluates a model's true positive rate against its false positive rate at various threshold settings. Among all tasks, the \textit{Profanity} task shows exhibits the best performance, recording a peak AUROC of 98.43. After a threshold of 0.5, the True Positive rate achieves its maximum and sustains this level, signifying consistently good generalization performance.

Figure~\ref{fig:fig11_right} illustrates the F1-score performance for the \textit{Profanity} task. After the 15 epochs, the F1-score remains stable, indicating that the model has converged in its learning for this task. The highest F1-score recorded is 95.84, showing the model's proficiency in accurately classifying profanities.

Figure~\ref{fig:fig12_left} represents the ROC curve encompassing all nine \textit{Bias} categories, providing a comprehensive view of the model's performance across diverse biases. The average AUROC across all biases stands at 94.34, indicating a generally high discriminative capability of the model. Within these categories, 'Race' achieves the highest value at 98.81, while 'Others' results in the lowest at 85.56 due to the relatively high difficulty of the task.

Figure~\ref{fig:fig13_right} illustrates the PRROC (Precision-Recall curve) for all the \textit{Bias} categories. This curve aids in understanding the model's precision and recall trade-offs across different biases. Notably, 'Region' stands out with the highest PRROC value of 95.02, demonstrating the model's strong capability to discern this particular bias. Meanwhile, 'Others' records the lowest PRROC of 49.58. The disparities in PRROC values across biases emphasize the importance of multi-metric evaluations to gain a comprehensive understanding of model performance.

\end{document}